\documentclass[conference]{IEEEtran}
\IEEEoverridecommandlockouts
\usepackage{graphicx}
\usepackage{amsmath,amsfonts,amssymb}
\usepackage[caption=false]{subfig}
\usepackage{bbm}
\usepackage{cite}

\def\BibTeX{{\rm B\kern-.05em{\sc i\kern-.025em b}\kern-.08em
    T\kern-.1667em\lower.7ex\hbox{E}\kern-.125emX}}

\begin{document}
\bstctlcite{IEEEexample:BSTcontrol}

\title{%
AutoQML: Automated Quantum Machine Learning \\for Wi-Fi Integrated Sensing and Communications}
              
\author{\IEEEauthorblockN{Toshiaki Koike-Akino, Pu Wang, Ye Wang}
\IEEEauthorblockA{
\textit{Mitsubishi Electric Research Laboratories (MERL), Cambridge, MA 02139, USA}
\\
Email: \{koike, pwang, yewang\}@merl.com
}
}

\maketitle

\begin{abstract}
Commercial Wi-Fi devices can be used for integrated sensing and communications (ISAC) to jointly exchange data and monitor indoor environment.
In this paper, we investigate a proof-of-concept approach using automated quantum machine learning (AutoQML) framework called \emph{AutoAnsatz} to recognize human gesture.
We address how to efficiently design quantum circuits to configure quantum neural networks (QNN).
The effectiveness of AutoQML is validated by an in-house experiment for human pose recognition, achieving state-of-the-art performance greater than 80\% accuracy for a limited data size with a significantly small number of trainable parameters. 
\end{abstract}

\begin{IEEEkeywords}
Integrated sensing and communication (ISAC), Wi-Fi sensing, human monitoring, quantum machine learning. 
\end{IEEEkeywords}

\section{Introduction}

A new paradigm called ``integrated sensing and communications (ISAC)'' has emerged as a key technology for future wireless systems~\cite{liu2022survey}.
In fact, many scenarios in the fifth generation and beyond (B5G), such as autonomous vehicles and extended reality, requires both high-performance sensing and data access.
In addition, millimeter wave (mmWave) and massive multiple-input multiple-output (MIMO) technologies used in B5G can achieve high resolution in both time and angular domain, bringing ISAC a viable concept.
In particular, Wi-Fi-based human monitoring has received much attention due to the decreasing cost and less privacy concerns compared with camera-based approaches. 
Modern deep neural networks (DNNs) have made Wi-Fi-band signals useful for user identification, emotion sensing, and skeleton tracking~\cite{AdibHsu15, HsuLiu17, ZhaoLi18, ZhaoTian18, ZhaoLu19, SinghSandha19, LuWen16, ZengPathak16, WuZhang17, ZouZhou18, YangZou18, GuLiu18, GuZhang19, WangJiang19, WangZhou19, WangFeng19, ma2019wifi, ZhangRuan20, JiangXue20, PajovicWang19, WangPajovic19, KoikeWang20,WangKoike20, YuWang20}.

In our recent work~\cite{koike2022quantum}, we introduced an emerging framework ``quantum machine learning (QML)''~\cite{henderson2020quanvolutional, romero2017quantum, rebentrost2014quantum, lloyd2018quantum, dallaire2018quantum, verdon2019quantum, huggins2019towards, cerezo2021cost, wang2021quantumnas, gomez2022towards, havlivcek2019supervised, schuld2020circuit, farhi2018classification, bergholm2018pennylane} into ISAC applications for the first time in literature, envisioning future era of \emph{quantum supremacy}~\cite{arute2019quantum, zhong2020quantum}.
QML is considered as a key driver in the sixth generation (6G) applications~\cite{nawaz2019quantum}, while there are few research yet to tackle practical problems.
Quantum computers have the potential to realize computationally efficient signal processing compared to traditional digital computers by exploiting quantum mechanism, e.g., superposition and entanglement, in terms of not only execution time but also energy consumption.
In the past few years, several vendors have successfully manufactured commercial quantum processing units (QPUs). 
For instance, IBM released $127$-qubit QPUs in 2021, and plans to produce $1121$-qubit QPUs by 2023.
It is no longer far future when noisy intermediate-scale quantum (NISQ) devices~\cite{bharti2021noisy} will be widely used for various real applications.
Recently, hybrid quantum-classical algorithms based on the \emph{variational} principle~\cite{farhi2014quantum, farhi2016quantum, anschuetz2019variational, kandala2017hardware} were proposed to deal with quantum noise for NISQ devices.

In this paper, we further extend our previous work towards automated QML (AutoQML)~\cite{gomez2022towards, wang2021quantumnas}, which we refer to as \emph{AutoAnsatz}, to facilitate quantum neural network (QNN) architecture design.
Most QNN models~\cite{bergholm2018pennylane} are based on particular quantum circuit templates known as \emph{ansatz}, which still requires careful tuning of hyperparameters such as the number of qubits, the number of entangling layers, etc. 
Finding suitable ansatz and hyperparameters generally involves a considerable amount of manual trial-and-error. 
The AutoQML~\cite{gomez2022towards, wang2021quantumnas} employs automated machine learning (AutoML) framework such as Bayesian optimization~\cite{akiba2019optuna} to automate QML design.
We experimentally validate the benefit of AutoQML in human pose recognition task using commercial-off-the-shelf (COTS) Wi-Fi devices.
The contributions of this paper are four-fold as described below.
\begin{itemize}
 \item This paper is the very first proof-of-concept study introducing AutoQML for Wi-Fi ISAC applications.
 \item We verify its feasibility for the human pose recognition application with COTS Wi-Fi devices. 
 \item We consider a diverse variety of ansatz to design QNN.
 \item We validate that the optimized QNN via AutoAnsatz achieves high accuracy comparable to state-of-the-art DNN while the QNN is configured with a significantly smaller number of trainable parameters.
\end{itemize}

\begin{figure}[t]
\centering
 \includegraphics[width=\linewidth]{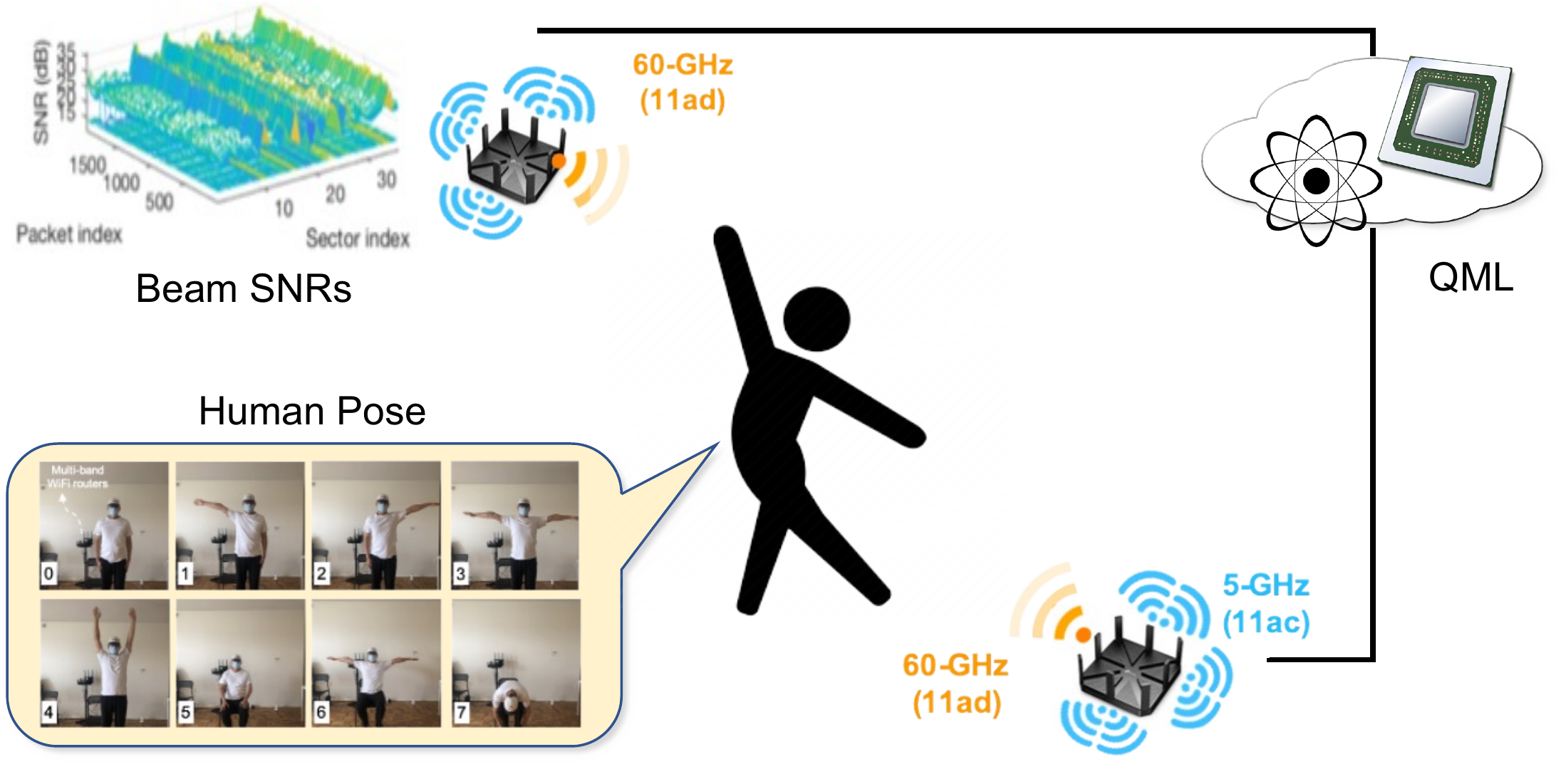}
 \caption{Wi-Fi sensing for human pose recognition empowered by QML.}
 \label{fig:system}
\end{figure}

\section{Quantum-Empowered Wi-Fi ISAC Systems}
\label{sec:algorithm}

Fig.~\ref{fig:system} shows the Wi-Fi human monitoring systems empowered by QML. 
During data communications, we collect beam scanning measurements associated with a class of human gestures as a fingerprinting data to learn QNN models.

\subsection{Human Pose Recognition Experiments}

Our experimental configuration uses two Wi-Fi stations: one station in front of a subject and another station behind the subject. 
Both stations are placed on a stand of a height of $1.2$~meters with a distance of approximately $2$ meters. 
The subject is asked to perform a total of $8$ poses including distinct gestures like `sit', `stand with left arm lifted', etc. 
For each pose, we recorded $7$ independent measurement sessions with different time duration and with sufficient time separation between consecutive two sessions. 
We use the measurements in the first four sessions as a training data and those in the last three sessions as a testing data. 
The total number of measurement samples is $42{,}915$ and $1{,}040$ in the training and testing, respectively.

\subsection{COTS Wi-Fi Testbed: mmWave Beam SNR}
As super-grained mmWave channel state information is not generally accessible from COTS devices without additional overhead, we use \emph{mid-grained} Wi-Fi measurements in the beam angle domain---beam signal-to-noise ratios (SNRs)---generated from the beam training (a.k.a. beam alignment) phase. 
For each probing beampattern (a.k.a. beam sectors), beam SNR is collected by $802.11$ad devices as a measure of beam quality.  
Such beam training is periodically carried out to adapt beam sectors to environmental changes. 
Accessing raw mmWave beam SNR measurements from COTS devices is enabled
via an open-source software~\cite{SteinmetzerWegemer18}. 

We use $802.11$ad-compliant TP-Link Talon AD7200 routers to collect beam SNRs at $60$ GHz.  
This router supports a single stream communication using analog beamforming over a $32$-element planar array. 
From one beam training, one Wi-Fi station can collect $36$ beam SNRs across discrete transmitting beampatterns. 
The measured beam SNRs are sent to a workstation via Ethernet cables to train a machine learning (ML) model over a cloud or on premise.
The experimental system is deployed in a standard indoor room setting. 
Further details of the experiments can be found in our previous work~\cite{YuWang20}.

\subsection{Quantum Machine Learning (QML)}

In~\cite{koike2022quantum}, we introduced QML framework to the Wi-Fi sensing systems, considering the rapid growth of quantum technology.
A number of modern DNN methods have been already migrated into quantum domain, e.g., convolutional layer~\cite{henderson2020quanvolutional}, autoencoder~\cite{romero2017quantum}, graph neural network~\cite{verdon2019quantum}, and generative adversarial network~\cite{lloyd2018quantum, dallaire2018quantum}.
Interestingly, the number of QML articles has been exponentially increasing at the same rate of DNN articles, doubling every year but just $6$ years behind.
It suggests that QML will be potentially used in numerous communities in a couple of years.
More importantly, QML is more suited for Wi-Fi sensing because a cloud quantum server such as IBM QX and Amazon braket is readily accessible.

In analogy with DNN, it was proved that QNN holds the universal approximation property~\cite{perez2020data}.
Accordingly, increasing the number of qubits may enjoy state-of-the-art DNN performance.
In addition, quantum circuits are analytically differentiable~\cite{schuld2019evaluating}, enabling stochastic gradient optimization of QNN.
Nevertheless, QNN often suffers from a vanishing gradient issue called the barren plateau~\cite{mcclean2018barren}.
To mitigate the issue, a simplified 2-design (S2D) ansatz~\cite{cerezo2021cost} was proposed to realize shallow entanglers for arbitrary unitary approximation.
It is highly expected that quantum computers would offer breakthroughs in a wide range of fields. 
As classical deep learning has become extremely energy intensive~\cite{strubell2019energy},
it is of importance to explore diverse computing modalities such as quantum computers for a future sustainable society.

\begin{figure}[t]
 \centering 
 \includegraphics[width=\linewidth]
 {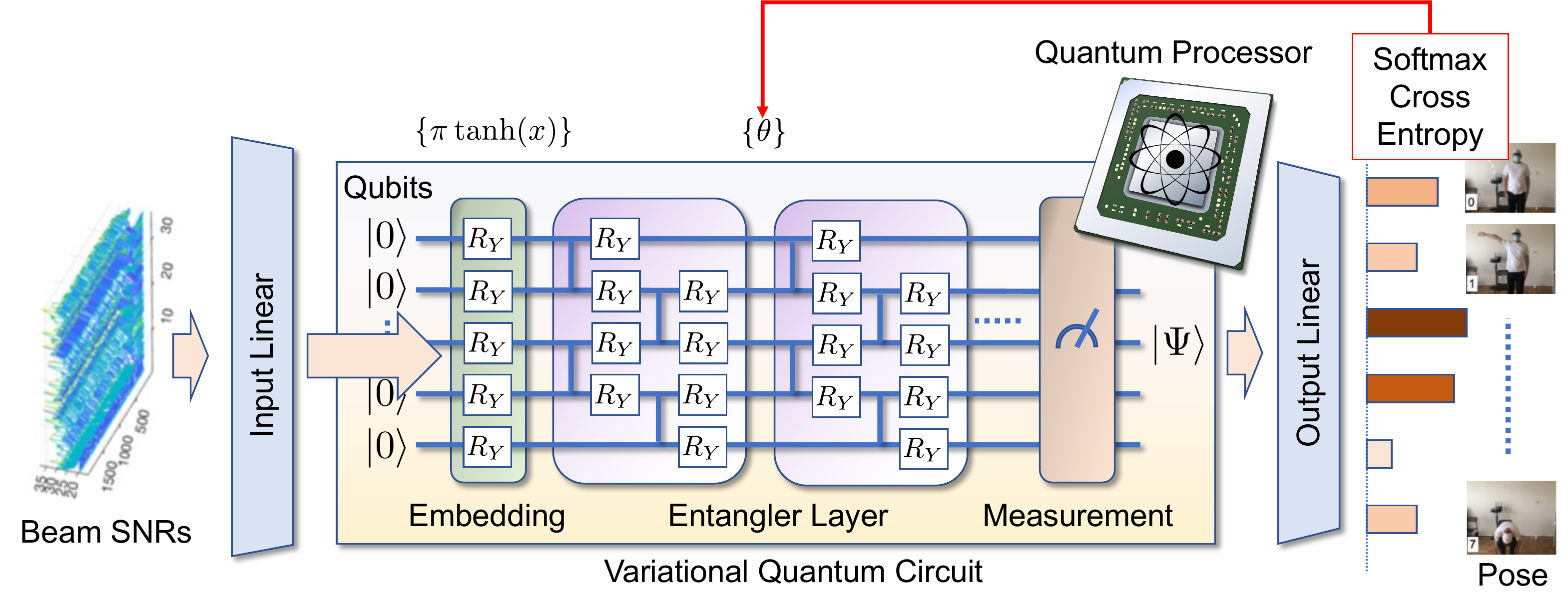}
 \caption{Variational QNN for pose recognition.}
 \label{fig:qnn}
\end{figure}

\subsection{Quantum Neural Network (QNN) for Wi-Fi Sensing}

Fig.~\ref{fig:qnn} depicts the QNN model used for Wi-Fi sensing in our previous paper~\cite{koike2022quantum}, employing S2D ansatz~\cite{cerezo2021cost}, which consists of Pauli-Y rotations and staggered controlled-Z entanglers.
This ansatz is a simplified variant of a $2$-design whose statistical properties are identical to ensemble random unitaries with respect to the Haar measure up to the first $2$ moments.
For an $n$-qubit variational quantum circuit, there are $2(n-1)L$ variational parameters $\{\theta\}$ over an $L$-layer S2D ansatz.

To feed $36$-dimensional beam SNRs, an input linear layer is used to initialize the quantum state for rotation angles of Pauli-Y gates.
The $8$-class pose estimation is provided by quantum measurements in the Hamiltonian observable of Pauli-Z operations, followed by an output layer to align the dimension.
The variational parameters as well as input/output layers are optimized by the adaptive momentum gradient method to minimize the softmax cross entropy loss.
While QNN is not necessarily better than DNN in prediction accuracy, it can be computationally efficient by manipulating $2^n$ quantum states in parallel with a small number of quantum gates.

\begin{figure}[t]
\centering
\subfloat[][Angle Embed]{
\includegraphics[width=0.23\linewidth]{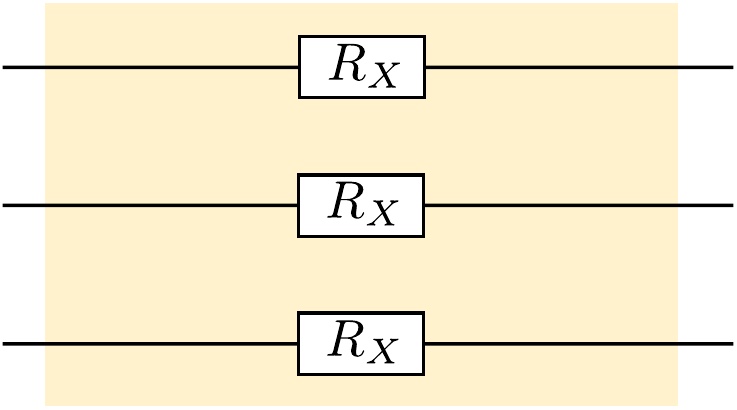}
\label{fig:angle}
}
\subfloat[][IQP Embed]{
\includegraphics[width=0.37\linewidth]{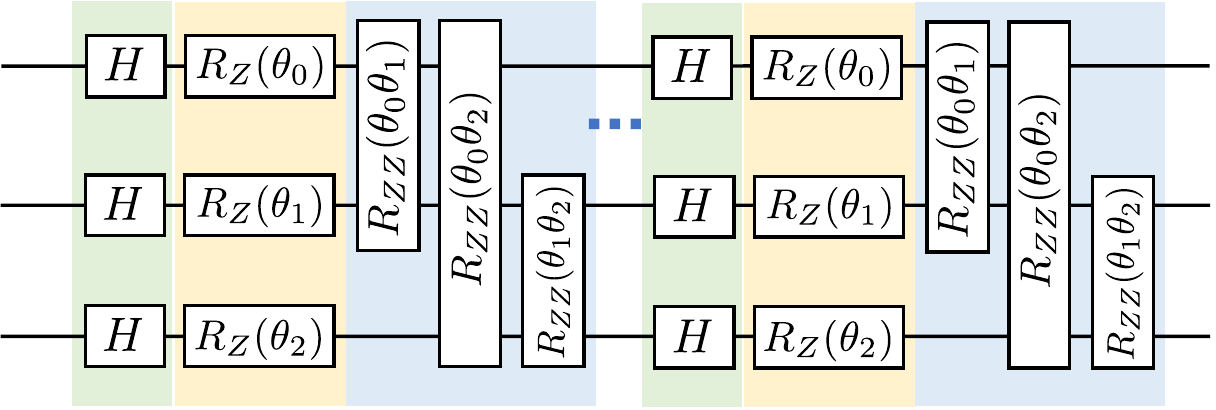}
\label{fig:iqp}
}
\subfloat[][Simple 2-Design]{
\includegraphics[width=0.33\linewidth]{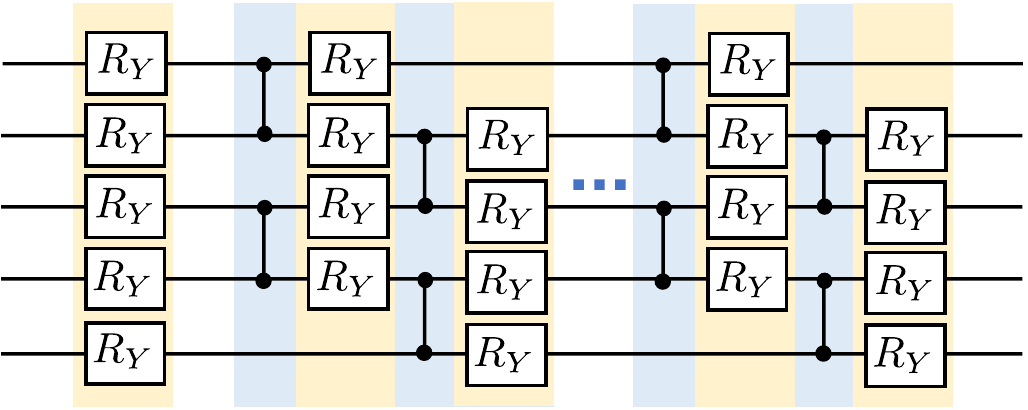}
\label{fig:s2d}
}
\\
\subfloat[][QAOA]{
\includegraphics[width=0.43\linewidth]{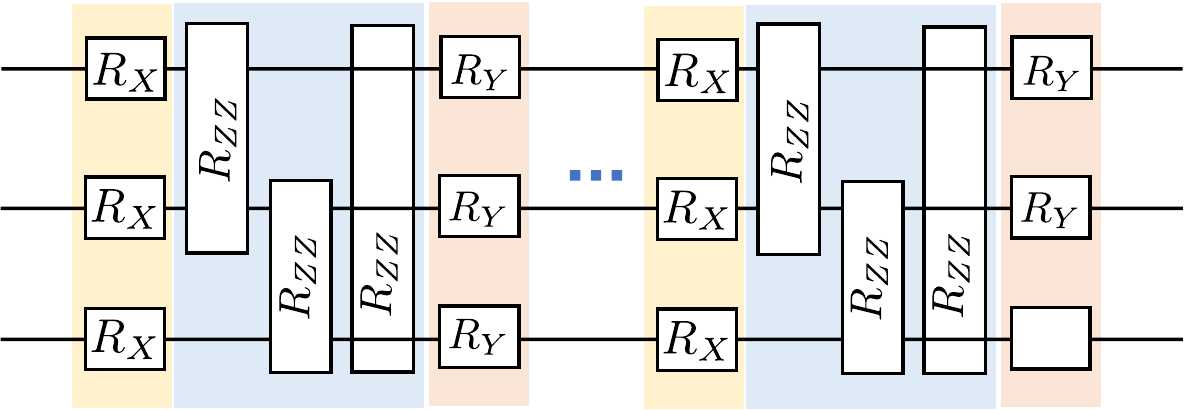}
\label{fig:qaoa}
}
\subfloat[][Tree Tensor Net]{
\includegraphics[width=0.24\linewidth]{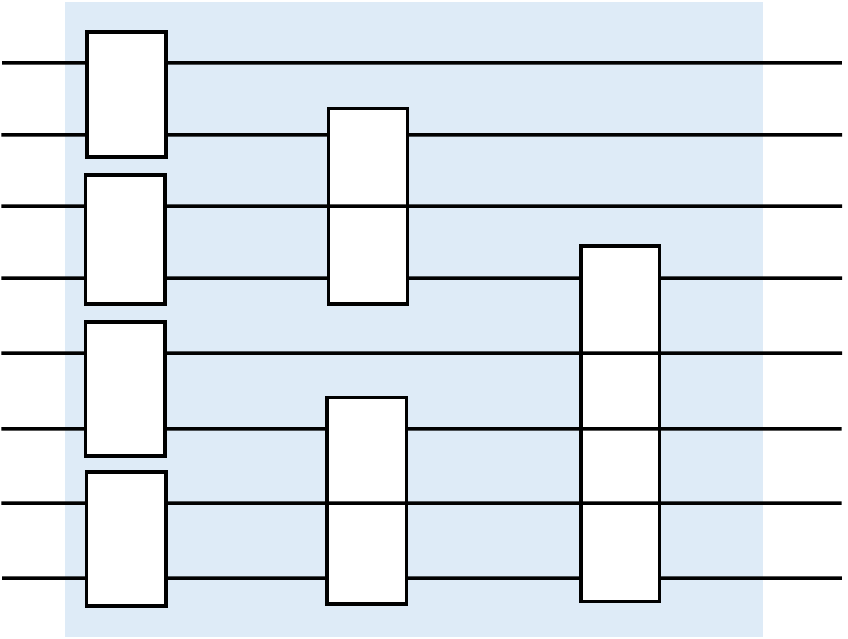}
\label{fig:ttn}
}
\subfloat[][Matrix Prod State]{
\includegraphics[width=0.24\linewidth]{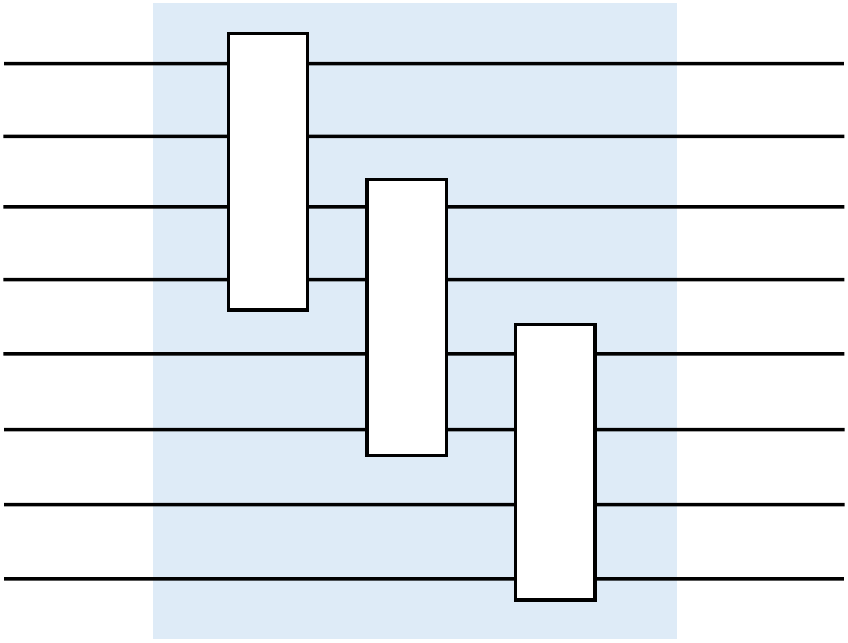}
\label{fig:mps}
}
\\
\subfloat[][Strong Entangle]{
\includegraphics[width=0.33\linewidth]{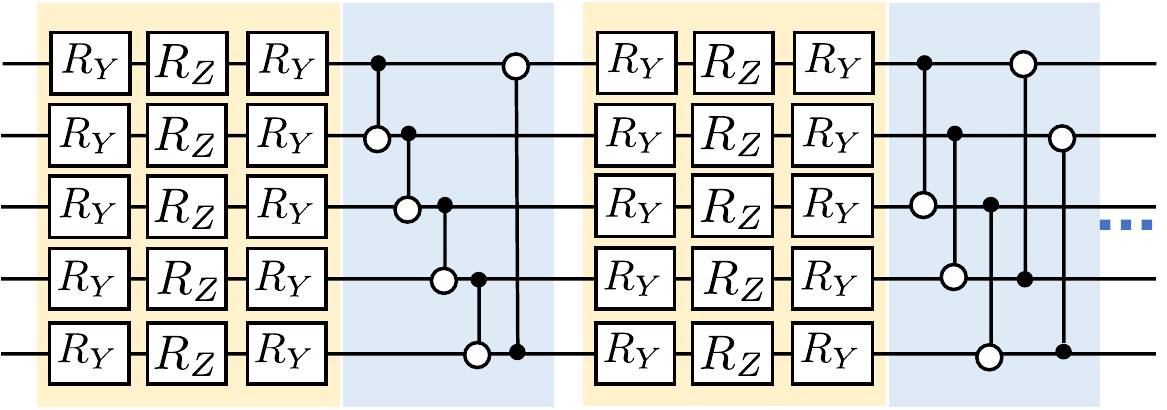}
\label{fig:se}
}
\subfloat[][Basic Entangle]{
\includegraphics[width=0.26\linewidth]{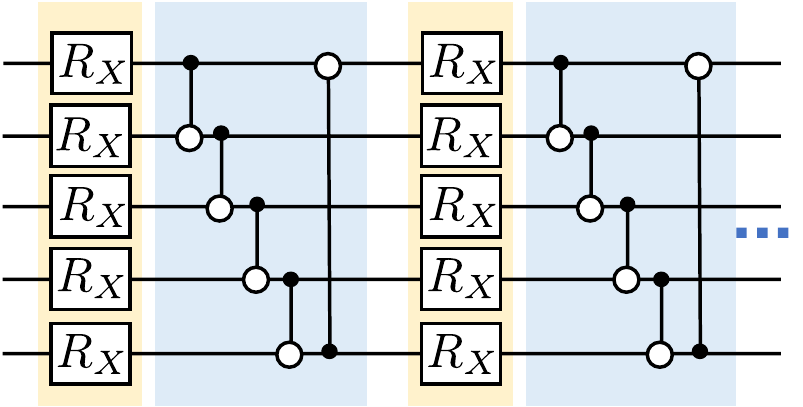}
\label{fig:be}
}
\subfloat[][Random]{
\includegraphics[width=0.33\linewidth]{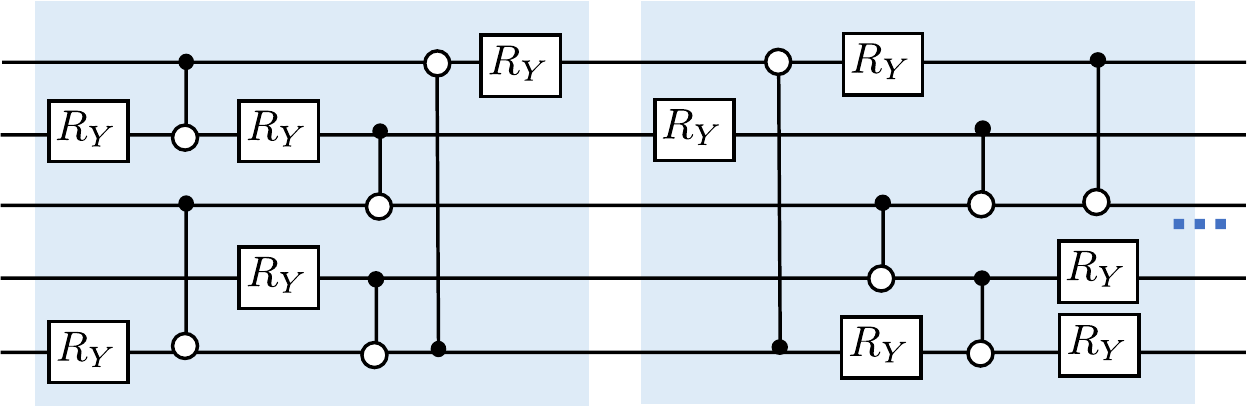}
\label{fig:r}
}
\caption{Quantum ansatz variety for QNN architecture.}
\label{fig:ansatz}
\end{figure}

\subsection{Automated QML for Ansatz/Hyperparameter Tuning}

The best QNN hyperparameters (e.g., the qubit size $n$ and layer size $L$) highly depend on datasets, and thus a considerable amount of manual effort is required to tune them in general.
In addition, there are various types of quantum ansatz in literature to explore besides S2D.
Fig.~\ref{fig:ansatz} illustrates some potential ansatz~\cite{bergholm2018pennylane}: angle embedding; instantaneous quantum polynomial time (IQP) embedding~\cite{havlivcek2019supervised}; quantum approximate optimization algorithm (QAOA)~\cite{farhi2014quantum}; tensor network~\cite{huggins2019towards} including tree tensor network (TTN) and matrix product state (MPS); basic entangler; strongly entangling layers; random layers.
In this paper, we propose \emph{AutoAnsatz} in the context of AutoQML to optimize the QNN ansatz and hyperparameters. 
Specifically, we use Bayesian optimization based on tree-Parzen estimator and hyperband pruning~\cite{akiba2019optuna}. 
It automatically explores the diverse set of quantum ansatz, qubit size, and layer size as well as learning rate, without the need of human effort in ansatz/hyperparameter tuning.

\section{Performance Evaluation}
\label{sec:performance}

\begin{figure}[t]
\centering 
\includegraphics[width=0.9\linewidth, trim={40 40 60 75}, clip]
{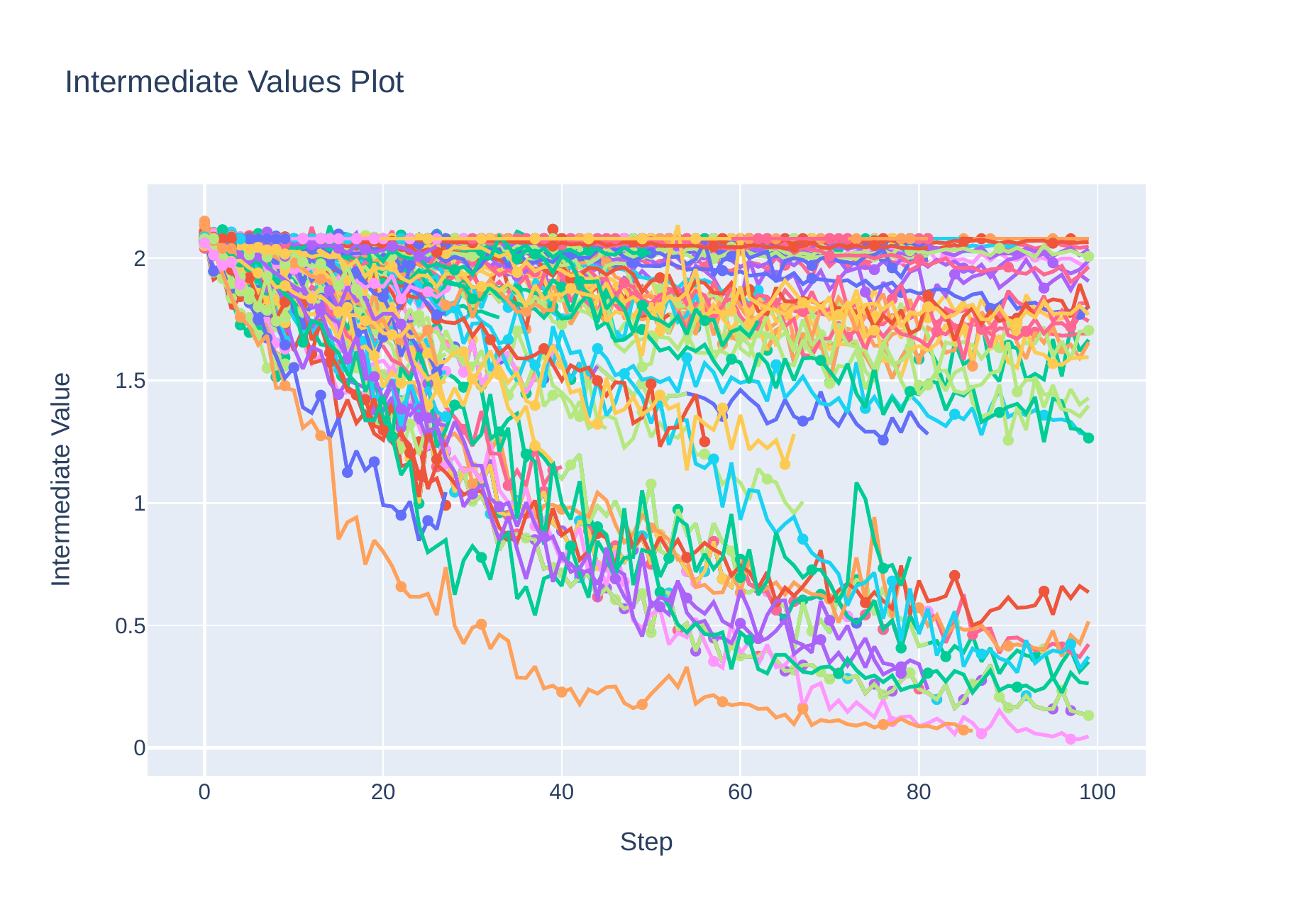}
\caption{Validation loss vs. learning epoch while AutoQML exploration with hyperband pruning ($0.3\%$ training data).}
\label{fig:pose_epoch}
\end{figure}

\begin{figure}[t]
\centering 
\includegraphics[width=\linewidth]
{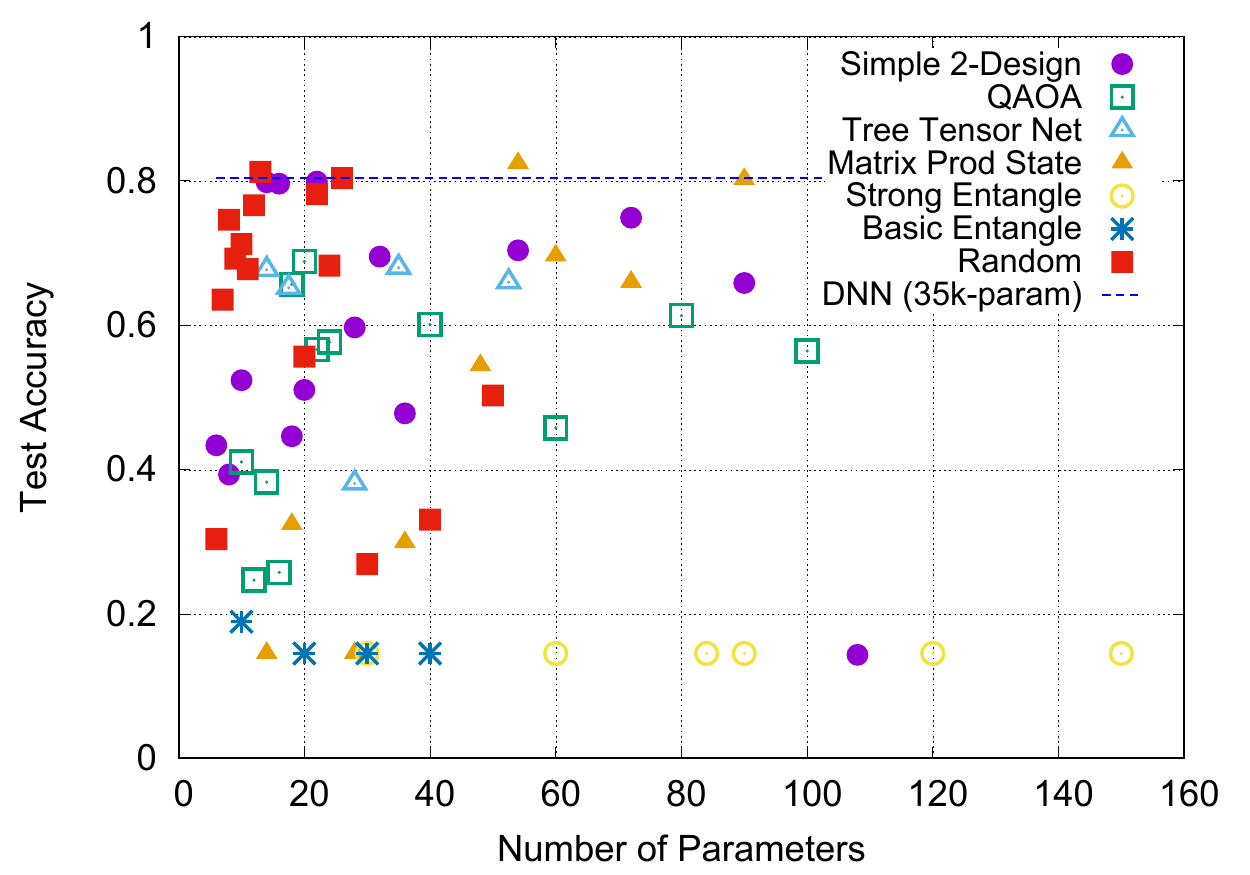}
\caption{Test accuracy vs. number of trainable parameters for various ansatz.}
\label{fig:pose_scatter}
\end{figure}

\subsection{AutoAnsatz Exploration Results}

AutoQML explores $2{,}000$ trials of hyperparameter tuning, where each model is trained with adaptive momentum (AdamW) gradient method over a maximum of $100$ epochs. 
The Bayesian optimization uses categorical sampling of seven different ansatz and two different embedding methods as shown in Fig.~\ref{fig:ansatz}.
The number of qubits $n$ and the number of entangler layers $L$ are also sampled from a range of $[5, \ldots, 15]$ and $[1, \ldots, 5]$, respectively.
In addition, the initial learning rate is optimized from a range of $(10^{-3}, 10^{-1})$, while the learning rate is adaptively decreased on plateau of training loss by a factor of $0.5$ over a patience of $10$ epochs.

Fig.~\ref{fig:pose_epoch} shows learning trajectories over training epoch, illustrating that some trials are efficiently pruned by the hyperband strategy due to hopeless or diverging trends. 
The pruning mechanism is important to accelerate automated ansatz search by early stopping useless trials.
One can see that there are many trials having extremely slow convergence.
The number of such pruned trials was about $89\%$ of total trials.

After AutoQML exploration, the best ansatz was chosen with $3$-layer MPS for $10$-qubit and angle embedding, having $54$ variational parameters.
Fig.~\ref{fig:pose_scatter} shows the test accuracy as a function of the total number of variational parameters for various models explored during the AutoQML.
It shows the tradeoff between performance and complexity.
The MPS ansatz tends to have better performance with more parameters, while most ansatz do not follow obvious trends.
It is partly because we constrained the number of learning epoch up to $100$, leading to poor underfitting performance for large-size QNN.
The best MPS model outperforms the state-of-the-art DNN model which has $650$-fold more trainable parameters (i.e., $35{,}000$).
We can also observe that S2D and random layers ansatz offer a good tradeoff between accuracy and complexity, achieving DNN performance at $2{,}000$-fold fewer parameters.

\begin{figure*}[t]
\centering 
\includegraphics[width=0.95\linewidth]
{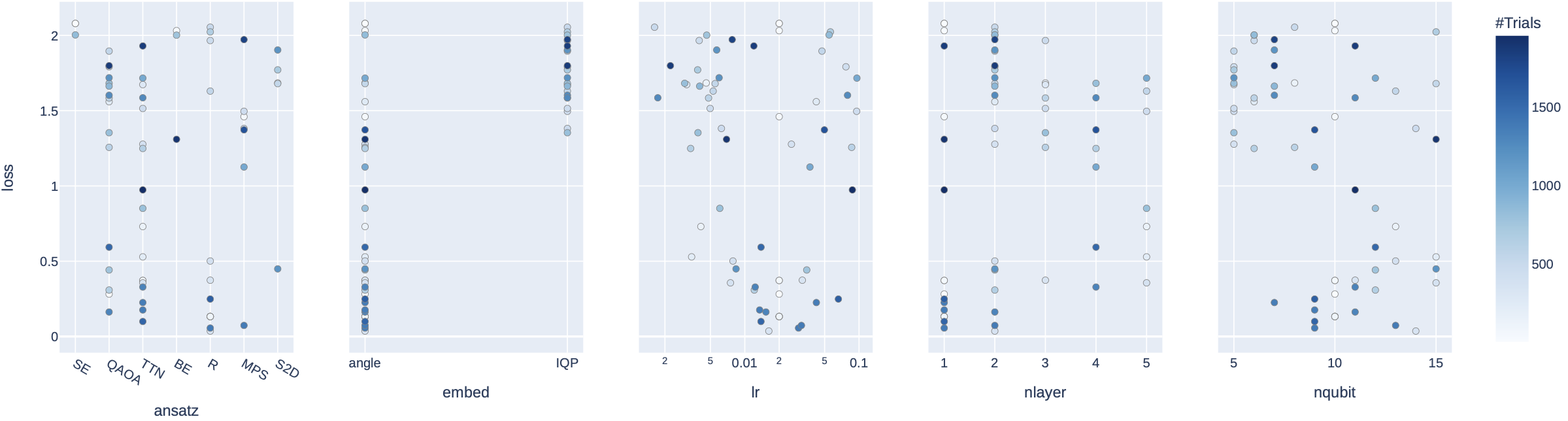}
\caption{Slice mapping of validation loss vs. single hyperparameter while AutoQML exploration.}
\label{fig:pose_slice}
\end{figure*}

\begin{figure}[t]
\centering 
\includegraphics[width=\linewidth, trim={0 10 10 70}, clip]
{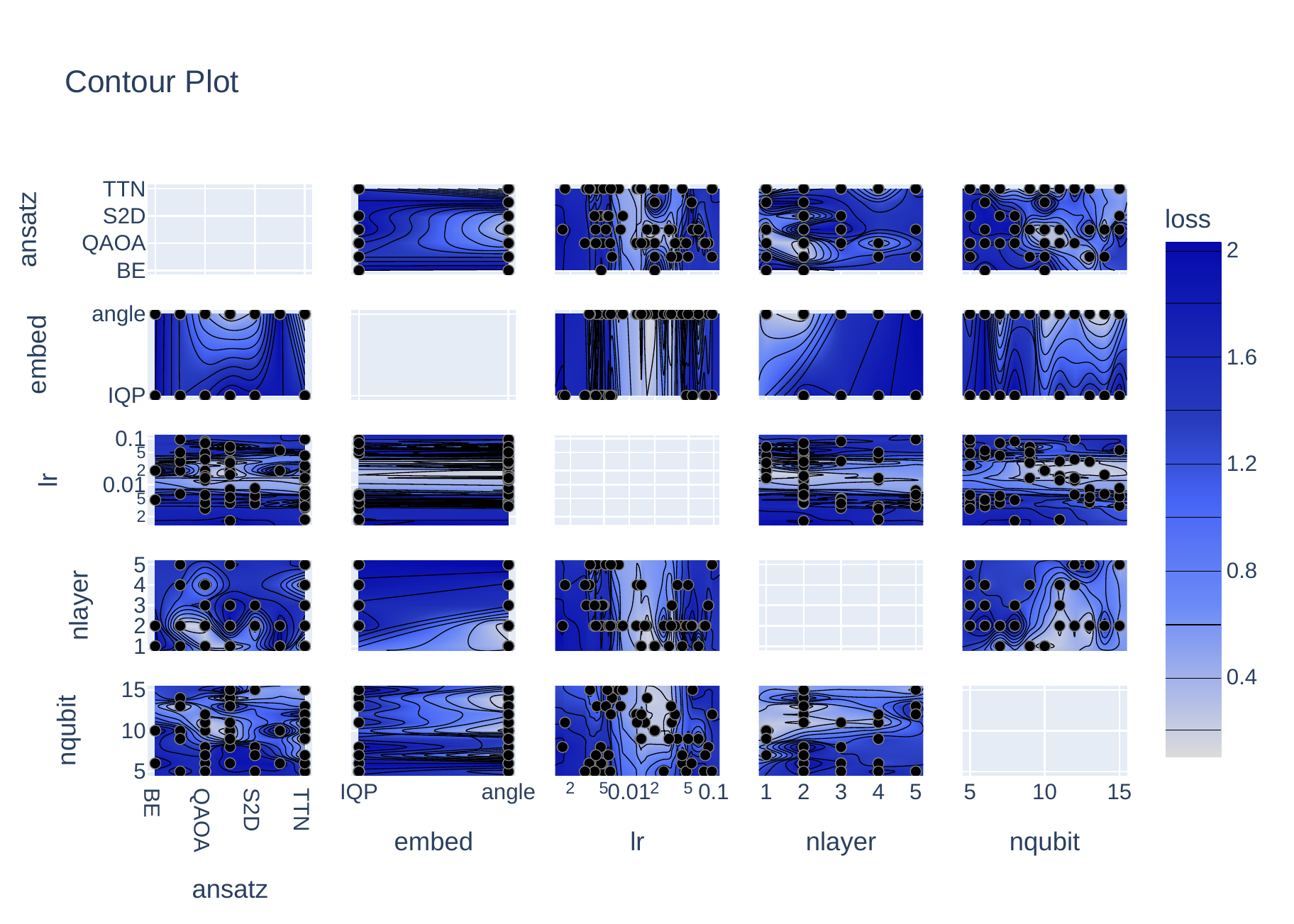}
\caption{Contour landscape of validation loss vs. two hyperparameters.}
\label{fig:pose_contour}
\end{figure}

\begin{figure}[t]
\centering 
\includegraphics[width=0.95\linewidth, trim={0 25 40 75}, clip]
{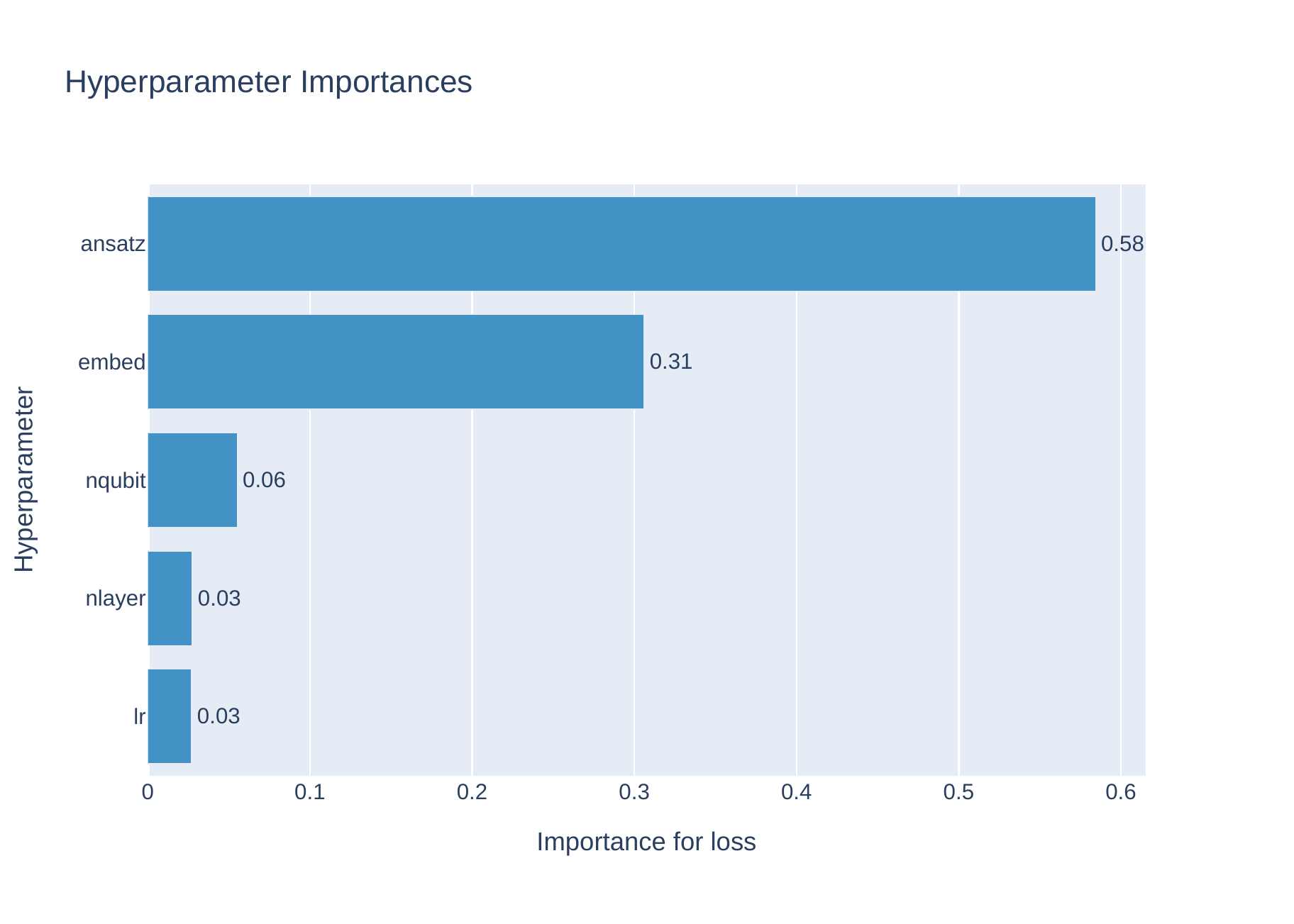}
\caption{fANOVA score on importance of QNN hyperparameters.}
\label{fig:pose_importance}
\end{figure}

\subsection{AutoAnsatz Analysis}

Fig.~\ref{fig:pose_slice} shows the relation between training loss and individual hyperparameter. 
For ansatz search, we can see that strongly entangling and basic entangler layers perform poorly. 
For embedding, IQP was not optimized well, partly because of complicated high-order ansatz for differentiation.
For learning rate, there is an obviously good choice at around $0.02$.
For the number of entangler layers, shallower QNNs tend to work better.
The number of qubits above $8$ (i.e., the number of pose classes) works equally well.

The loss contour is plotted in Fig.~\ref{fig:pose_contour} showing a non-trivial landscape for the combination of two hyperparameters, indicating difficulty for manual tuning. 
Fig.~\ref{fig:pose_importance} presents the functional analysis of variance (fANOVA) score~\cite{akiba2019optuna} to assess the importance of hyperparameters.
It is found that the choice of ansatz is the most influential hyperparameter, while the learning rate (which is usually most important for DNN design) is the lowest important one.

\subsection{Comparison of Machine Learning (ML) Models}

Finally, we compare different ML models including support vector machine (SVM), decision tree (DT), $k$-nearest neighbor ($k$NN), Gaussian na\"{i}ve Bayes (GNB), random forest (RF), and extra tree ensemble methods ($10$ base models).
For baseline DNN, we consider residual $4$ hidden layers with $100$ hidden nodes using Mish activation, with approximately $35$k trainable parameters.
For baseline QNN, we use $10$-qubit $1$-layer S2D ansatz~\cite{koike2022quantum}, having $18$ quantum variational parameters.
We use AdamW optimizer for a mini batch size of $100$ over $100$ epochs with a learning rate of $0.02$ and weight decay of $10^{-4}$.

\begin{figure}[t]
\centering 
\includegraphics[width=\linewidth]
{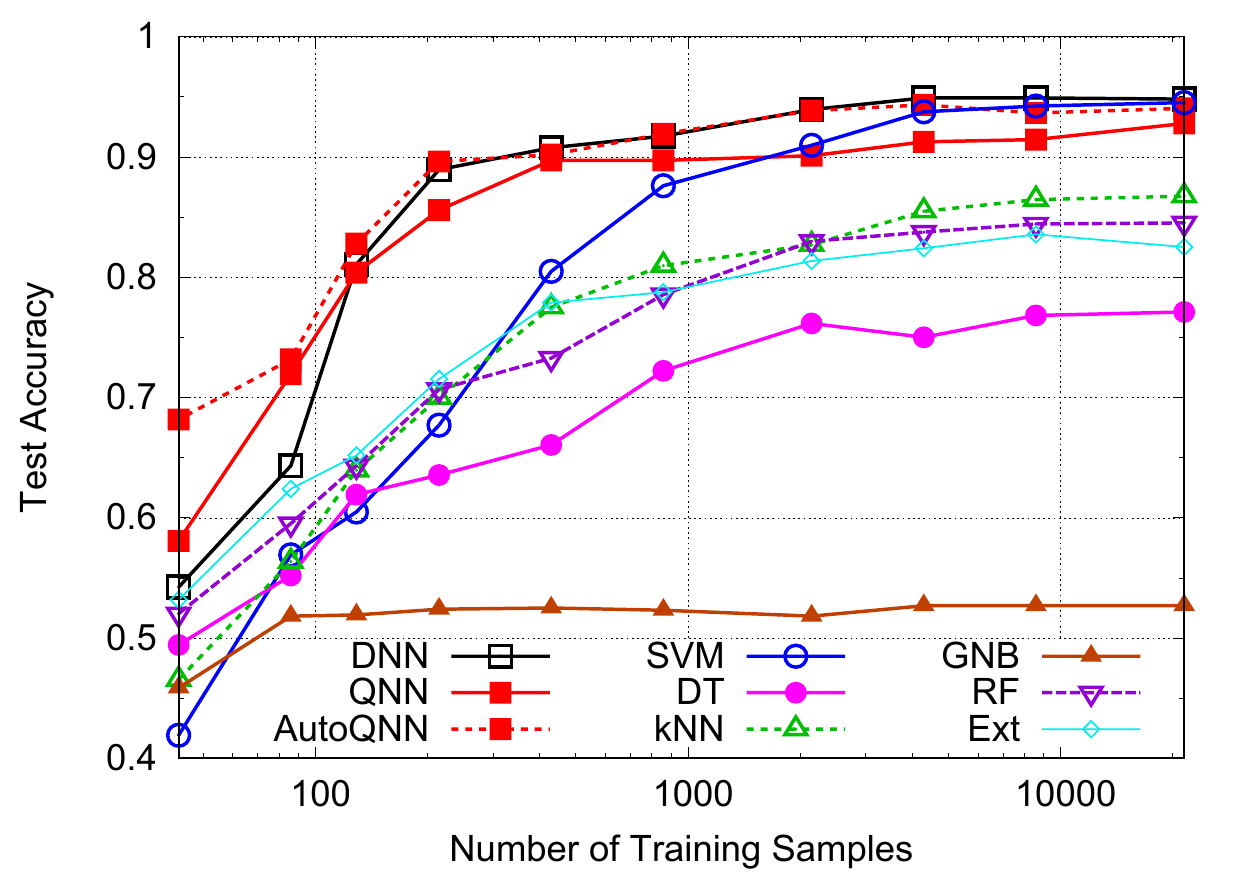}
\caption{Test accuracy vs. training samples for various ML methods.}
\label{fig:pose_train}
\end{figure}

Fig.~\ref{fig:pose_train} shows the test accuracy as a function of the number of labeled training samples.
One can notice that the performance can exceed an accuracy of $90\%$ for DNN, QNN, and SVM when a sufficient amount of labeled data is available.
It is confirmed that a small-scale QNN designed by AutoQML can improve the baseline QNN model, achieving state-of-the-art performance comparable to a large-scale DNN.

\section{Conclusion}
\label{sec:conclusion}
We investigated AutoQML to design ansatz of QNN for Wi-Fi sensing tasks.
We demonstrated the benefit of AutoQML/AutoAnsatz through real-world experiments with an in-house Wi-Fi testbed. 
It was shown that a small-scale QNN can achieve state-of-the-art performance comparable to a large-scale DNN, in human pose recognition. 
Validation on real quantum processors will be provided in a future work.
This is a very initial proof-of-concept study for quantum-ready Wi-Fi sensing and there remain many fascinating open issues.

\section*{Acknowledgment}
We thank Jianyuan Yu (Virginia Tech.) for earlier data collection experiments.

\bibliographystyle{IEEEtran}
\bibliography{references, bib_localization, quantum.bib}

\end{document}